\documentclass[runningheads]{llncs}
\usepackage{graphicx}
\usepackage{cite} 
\usepackage{times}
\usepackage{epsfig}
\usepackage{amsmath}
\usepackage{amssymb}
\usepackage{mwe}
\usepackage{acro}
\usepackage{amssymb}
\usepackage{xcolor,colortbl}
\usepackage{tabularx}
\usepackage{relsize}
\usepackage{pifont}
\usepackage{booktabs} 
\usepackage{multirow}
\usepackage{multicol}
\usepackage{adjustbox}
\usepackage{float}
\usepackage{xcolor}

\usepackage[hidelinks]{hyperref}
\hypersetup{
  colorlinks   = true, 
  urlcolor     = red, 
  linkcolor    = blue, 
  citecolor   = green 
}

\DeclareAcronym{ROI}{
short=ROI,
long=region of interest,
}

\DeclareAcronym{IOU}{
short=IOU,
long=intersection over union,
}

\DeclareAcronym{cIOU}{
short=cIOU,
long=circle intersection over union,
}

\DeclareAcronym{DoF}{
short=DoF,
long=degrees of freedom,
}

\newcommand{\etal}{\textit{et al}. }

\newcommand{\Fig}{Fig.}
\begin{document}
\title{CircleSnake: Instance Segmentation with Circle Representation}
%
%

\author{Ethan H. Nguyen\inst{1} \and 
Haichun Yang\inst{2} \and
Zuhayr Asad\inst{1} \and
Ruining Deng\inst{1} \and
Agnes B. Fogo\inst{2} \and
Yuankai Huo\inst{1}}

\institute{Vanderbilt University , Nashville TN 37235, USA \and
Vanderbilt University Medical Center, Nashville TN 37215, USA 
\email{yuankai.huo@vanderbilt.edu}}

\maketitle              
\begin{abstract}
Circle representation has recently been introduced as a ``medical imaging optimized" representation for more effective instance object detection on ball-shaped medical objects. With its superior performance on instance detection, it is appealing to extend the circle representation to instance medical object segmentation. In this work, we propose CircleSnake, a simple end-to-end circle contour deformation-based segmentation method for ball-shaped medical objects. Compared to the prevalent DeepSnake method, our contribution is threefold: (1) We replace the complicated \textit{bounding box to octagon contour} transformation with a computation-free and consistent \textit{bounding circle to circle contour} adaption for segmenting ball-shaped medical objects; (2) Circle representation has fewer degrees of freedom (DoF=2) as compared with the octagon representation (DoF=8), thus yielding a more robust segmentation performance and better rotation consistency; (3) To the best of our knowledge, the proposed CircleSnake method is the first end-to-end circle representation deep segmentation pipeline  method with consistent circle detection, circle contour proposal, and circular convolution. The key innovation is to integrate the circular graph convolution with circle detection into an end-to-end instance segmentation framework, enabled by the proposed simple and consistent circle contour representation. Glomeruli are used to evaluate the performance of the benchmarks. From the results, CircleSnake increases the average precision of glomerular detection from 0.559 to 0.614. The Dice score increased from 0.804 to 0.849. The code has been released: {\color{red}{https://github.com/hrlblab/CircleSnake}}

\end{abstract}

\begin{figure}[t]
\begin{center}
\includegraphics[width=1\linewidth]{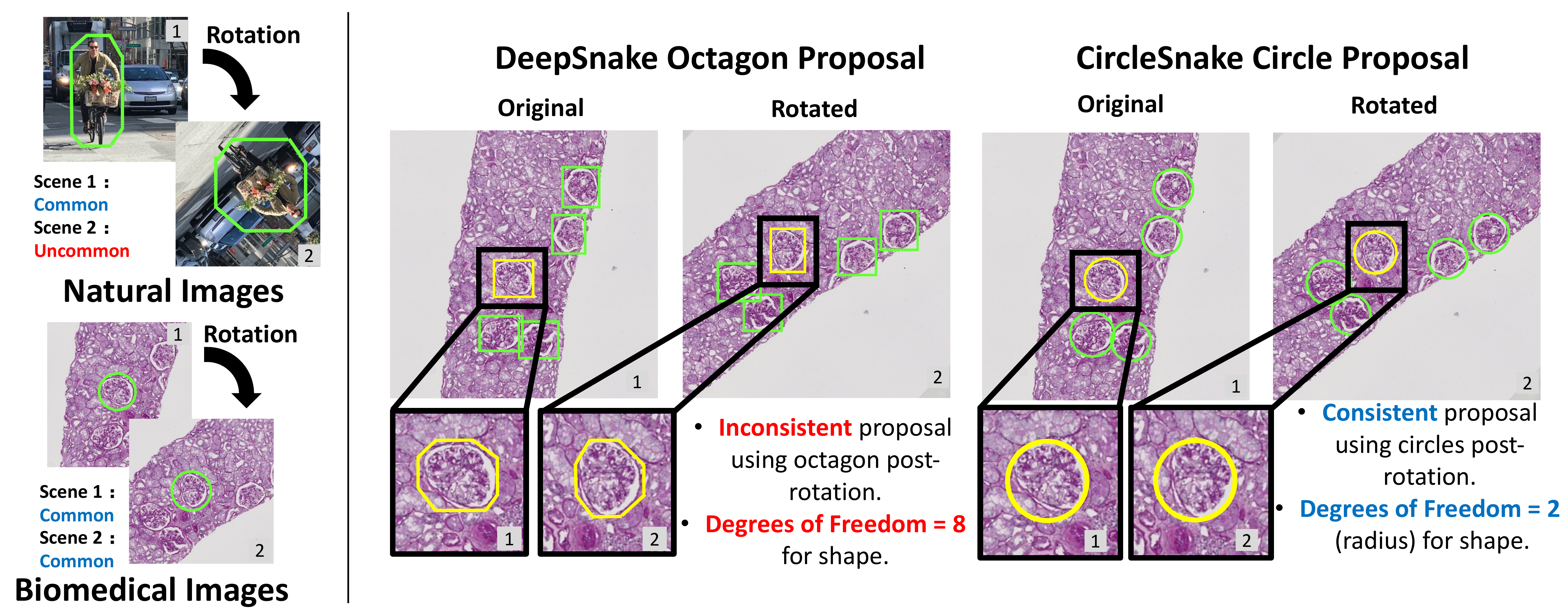}
\end{center}
   \caption{\textbf{Comparison of traditional representation and the circle representation.} The left panel shows that samples of glomeruli can be scanned at any angle of rotation. The right panel highlights how the octagon proposal is not optimized for the ball-shaped glomerulus. Using the proposed CircleSnake yields a more consistent representation while using fewer degrees of freedom (DoF). Moreover, the proposed circle contour proposal is fundamentally easier and more consistent as compared with the computer vision-oriented octagon contour proposal (\Fig~\ref{fig2}).}
\label{fig1}
\end{figure}

\section{Introduction}
Bounding box and polygon representations from the computer vision community have been widely utilized for medical object detection~\cite{lo2018glomerulus, kawazoe2018faster, heckenauer2020real, rehem2021automatic}. However, such representations are not necessarily optimized (\Fig~\ref{fig1}) for medical objects. Unlike natural images, certain biomedical images (e.g., microscopy imaging can be obtained and displayed at any angle of rotation of the same tissue~\cite{yang2020circlenet}. Thus, traditional object representation might yield inferior performance for representing such ball-shaped objects~\cite{nguyen2021circle}.

\begin{figure}[t]
\begin{center}
\includegraphics[width=0.85\linewidth]{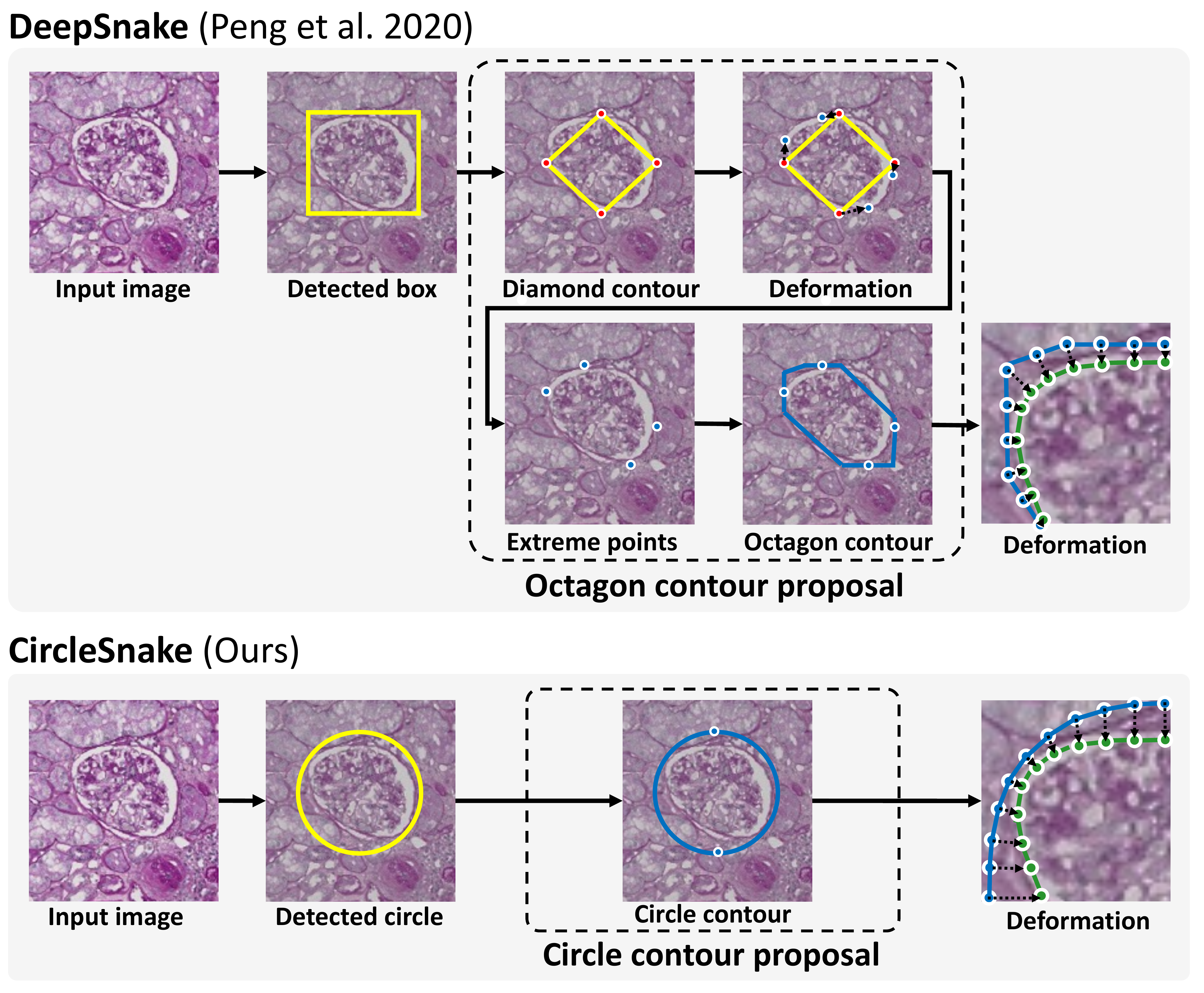}
\end{center}
   \caption{\textbf{Circle contour proposal.} This figure presents the differences between the ``bounding box to octagon contour" representation and the proposed ``bounding circle to circle contour" representation. Our circle contour proposal avoids the complicated extreme points and deformation-based contour proposal by directly introducing a simple circle proposal, which bridges the circle detection and deformation-based segmentation in an computational-free and end-to-end manner.}
\label{fig2}
\end{figure}

Recently, circle representation has been introduced as a ``medical imaging optimized" representation for instance object detection on ball-shaped medical objects, such as glomeruli, nuclei, and tumors~\cite{yang2020circlenet, nguyen2021circle, luo2021scpmnet}. Given its superior performance, it is appealing to extend the circle representation to the instance object segmentation. In renal pathology, the ability to precisely segment glomeruli is critical to investigating several kidney diseases~\cite{d2013rise, huo2021ai}. Most prior arts~\cite{gadermayr2017cnn,bueno2020glomerulosclerosis,kannan2019segmentation,ginley2019computational} are pixel-based where they perform instance segmentation within a regional proposal on the pixel level. A popular example, Mask R-CNN~\cite{he2017mask}, first detects objects and then segments instances within the proposed boxes using a mask predictor. In contrast, DeepSnake~\cite{peng2020deep} addresses localization errors from the detector by deforming the detected bounding octagon to object boundaries. However, DeepSnake's bounding octagon is not optimal for the ball-shaped glomerulus (\Fig~\ref{fig1}) and furthermore, is computationally complicated (\Fig~\ref{fig2}).

In this paper, we propose a contour-based approach that utilizes the circle representation called CircleSnake for the robust segmentation of glomeruli. The ``bounding circle" is introduced as the detection and initial contour representation for the ball-shaped structure. Once the center location of the lesion is obtained, only \ac{DoF} = 2 is required to form the bounding circle, while \ac{DoF} = 8 is required for the bounding octagon. Briefly, the contributions of this study are in three areas:

$\bullet$ \textbf{Optimized Biomedical Object Segmentation}:  To the best of our knowledge, the proposed CircleSnake is the first contour deformation-based end-to-end segmentation approach for detecting ball-shaped medical objects.

$\bullet$ \textbf{Circle Representation}: We propose a consistent circle representation pipeline for segmenting ball-shaped biomedical objects with integrated (1) circle detection, (2) circle contour proposal, and (3) circular convolution, with a smaller \ac{DoF} of fitting and superior segmentation performance.

$\bullet$ \textbf{Rotation Consistency}: The proposed circle  representation yields fewer \ac{DoF} of fitting, better segmentation performance, and superior rotation consistency.

\begin{figure}
\begin{center}
\includegraphics[width=1\linewidth]{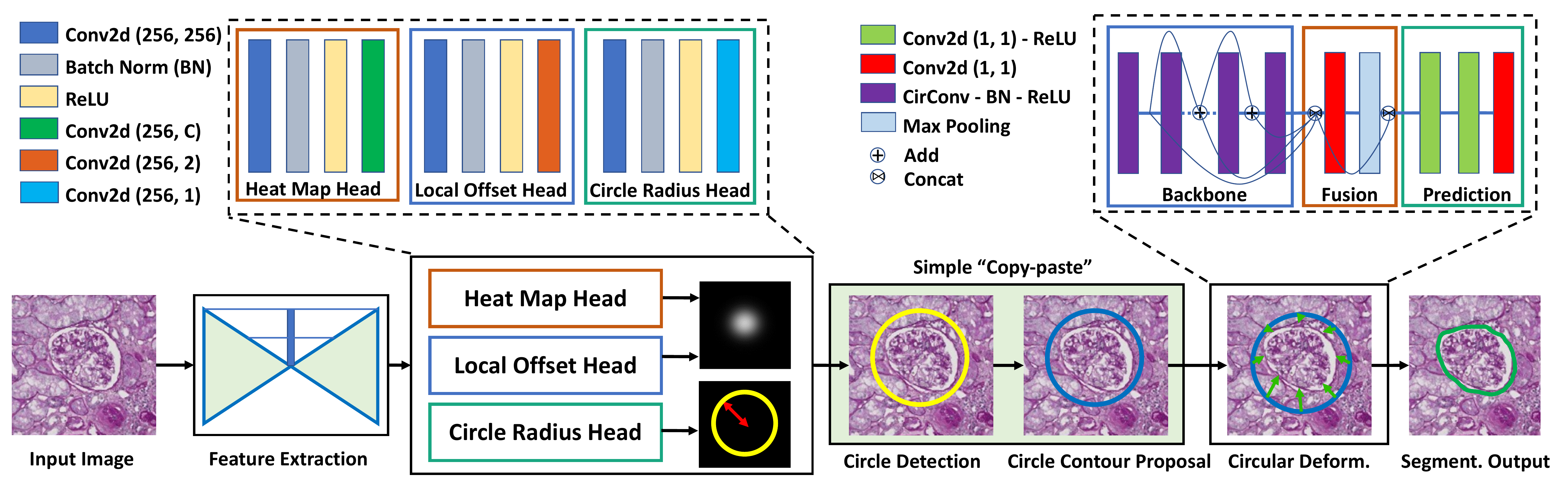}
\end{center}
   \caption{\textbf{The network structure of proposed CircleSnake}. The proposed CircleSnake After detection by CircleNet, the bounding circle forms the initial contour. This contour is deformed into the final contour using a graph convolutional network (GCN).}
\label{fig:network}
 \end{figure}

\section{Methods}
\Fig~\ref{fig:network} provides an overview of the proposed pipeline for instance segmentation. Inspired by~\cite{peng2020deep}, the object segmentation is implemented by deforming an initial rough contour to match object boundary. Different from DeepSnake which uses bounding box representation and polygon contour, we propose to use circle detection~\cite{nguyen2021circle} and circle contour proposal (\Fig~\ref{fig1}) in our CircleSnake method. Then the proposed circle initial contour is deformed via circular convolution by predicting the per-vertex offsets towards the real object boundary (\Fig~\ref{fig2}). Such deformation is performed iteratively to obtain the final object shape as instance segmentation.

\subsection{Circle Object Detection}
The circle object detection follows the design of the recent CircleNet approach~\cite{nguyen2021circle,yang2020circlenet} for its high performance and simplicity (Fig.~\ref{fig:network}). Following the definitions from Zhou \etal~\cite{zhou2019objects}, let $I \in R^{W \times H \times 3}$ be an input image with height $H$ and width $W$. The CPL network produces the center localization of each object within a heatmap $\hat Y \in [0,1]^{\frac{W}{R} \times \frac{H}{R} \times C}$ where $C$ is the number of candidate class and $R$ is a downsampling factor. Within the heatmap, $\hat Y_{xyc} = 1$ is the center of the lesion and $\hat Y_{xyc} = 0$ is the background. Following convention, ~\cite{law2018cornernet, zhou2019objects}, the target center point is splat on a heatmap as a 2D Gaussian kernel:
\begin{equation}
{Y_{xyc} = \exp\left(-\frac{(x-\tilde p_x)^2+(y-\tilde p_y)^2}{2\sigma_p^2}\right)}
\end{equation} 
where the $\tilde p_x$ and $\tilde p_y$ are the downsampled ground truth center points and $\sigma_p$ is the kernel standard deviation. The training loss $L_{k}$ is penalty-reduced logistic regression with focal loss~\cite{lin2017focal}:
\begin{equation}
    L_k = \frac{-1}{N} \sum_{xyc}
    \begin{cases}
        (1 - \hat{Y}_{xyc})^{\alpha} 
        \log(\hat{Y}_{xyc}) & \!\text{if}\ Y_{xyc}=1\\
        \begin{array}{c}
        (1-Y_{xyc})^{\beta} 
        (\hat{Y}_{xyc})^{\alpha}\\
        \log(1-\hat{Y}_{xyc})
        \end{array}
        & \!\text{otherwise}
    \end{cases}
\end{equation}
where the hyper-parameters $\alpha$ and $\beta$ are kept the same as~\cite{lin2017focal}. To further refine the prediction location, the $\ell_1$-norm offset prediction loss $L_{off}$ is used.

To obtain the center point, the top $n$ peaks whose values are greater or equal to its 8-connected neighbors are proposed.
These $n$ detected center points are defined as $\hat{\mathcal{P}} = \{(\hat x_i, \hat y_i)\}_{i = 1}^{n}$. The center point of each object contains an integer coordinate $(x_i,y_i)$ from $\hat Y_{x_iy_ic}$ and $L_{k}$. The offset $(\delta \hat x_i, \delta \hat y_i)$ is obtained from $L_{off}$. The bounding circle is combines the center point $\hat{p}$ and radius $\hat{r}$ as:
\begin{equation}
 \hat{p} = (\hat x_i + \delta \hat x_i ,\ \ \hat y_i + \delta \hat y_i). \quad \hat{r} = \hat R_{\hat x_i,\hat y_i}.
\end{equation}
where $\hat R  \in \mathcal{R}^{\frac{W}{R} \times \frac{H}{R} \times 1}$ contains the radius prediction for each pixel, optimized by 
\begin{equation}
    L_{radius} = \frac{1}{N}\sum_{k=1}^{N} \left|\hat R_{p_k} - r_k\right|.
\end{equation}
where $r_k$ is the ground truth radius for each object $k$. Finally, the overall objective is
\begin{equation}
    L_{det} = L_{k} + \lambda_{radius} L_{radius} + \lambda_{off}L_{off}.
\label{eq:total_loss}
\end{equation} Following ~\cite{zhou2019objects}, we set $\lambda_{radius} = 0.1$ and $ \lambda_{off} = 1$.

\subsection{Circle Contour Proposal}
Using the CircleNet based object detection~\cite{yang2020circlenet}, bounding circle based object detection is obtained for every target object. Then, the initial circle contour proposal is directly achieved from the circle representation (\Fig~\ref{fig2}). Our new circle contour proposal skips the the original complicated deformation and extreme point-based octagon contour proposal; ultimately, this makes the contour proposal both simple and consistent. The circle proposal is determined by the center point and radius. $N$ initial points $\{\mathbf{x}^{circle}_i | i = 1, 2,... , N\}$ are uniformly sampled from the circle contour starting at the top-most point $x_{1}^{circle}$ . Similarly, the ground truth contour is formed by sampling $N$ vertices clockwise along the object boundary. The $N$ is set to 128 based on~\cite{peng2020deep}.

\subsection{Circular Contour Deformation}
Given a contour with $N$ vertices $\{\mathbf{x}^{circle}_i | i = 1, ..., N\}$, we first construct feature vectors for each vertex. The input feature $f^{circle}_i$ for a vertex $\mathbf{x}^{circle}_i$ is a concatenation of learning-based features and the vertex coordinate: $[F(\mathbf{x}^{circle}_i); \mathbf{x}^{circle}_i]$, where $F$ denotes the feature maps. The input features defined are treated as a 1-D discrete signal $f: \mathbb{Z} \to \mathbb{R}^D$ on a circle contour. We utilize the circular convolution for the feature learning, as illustrated~\cite{peng2020deep}. The $f$ is defined as a periodic signal as:
\begin{equation}
    (f_N^{circle} \ast k)_i = \sum_{j = -r}^r (f_N^{circle})_{i + j}k_j,
\end{equation}
where $k: [-r, r] \to \mathbb{R}^D$ is a learnable kernel function, while the operator $\ast$ is the standard convolution. Following~\cite{peng2020deep}, the kernel size of the circular convolution is fixed to be nine.

In CircleSnake, the above convolution is implemented via the graph convolutional network (GCN) inspired by~\cite{peng2020deep}. The GCN consists of three parts: backbone, fusion, and prediction. The backbone contains eight "CirConv-Bn-ReLU" layers with residual skip connections and "CirConv" for circular convolution. The fusion block combines the information across contour points at different scales. Specifically, the features are concatenated from all layers in the backbone and forwarded through a 1$\times$1 convolutional layer and max pooling layer. Finally, the prediction head contains three 1$\times$1 convolution layers and outputs vertex-wise offsets. The loss function for the iterative contour deformation is defined as
\begin{equation}
    L_{iter} = \frac{1}{N}\sum_{i=1}^{N} l_{1}(\tilde x_{i}^{circle} - x_{i}^{gt}).
\end{equation}
where $x_{i}^{gt}$ is the ground truth boundary point and $\tilde x_{i}^{circle}$ is the deformed contour point. Following~\cite{peng2020deep}, we regress the $N$ offsets in 3 iterations. 

\section{Experimental Design}

For baseline methods, DeepSnake~\cite{peng2020deep} was utilized for their superior performance in instance segmentation. Deep layer aggregation (DLA) network ~\cite{yu2018deep} were employed as backbone networks. The implementations the backbone networks and segmentation networks followed the authors' official PyTorch implementations. All models were initialized with the COCO pretrained model ~\cite{lin2014microsoft}. The same workstation with an Nvidia 3090 was used to perform all experiments in this study. For both the glomeruli and nuclei experiments, the hyperparameters were set to maximum epoch = 50, learning rate = $1e-4$, batch size = 16, and optimizer = Adam.

Conventional segmentation metrics were used. For detection, we utilized, average precision ($AP$), $AP_{50}$ (IOU threshold at 0.5), $AP_{75}$ (IOU threshold at 0.75), $AP_S$ (small scale with area $<$ $32^2$), $AP_M$ (medium scale with area $>$ $32^2$). For segmentation, Dice score was used. as~\cite{nguyen2021circle}.

\begin{figure}[t]
\begin{center}
\includegraphics[width=1\linewidth]{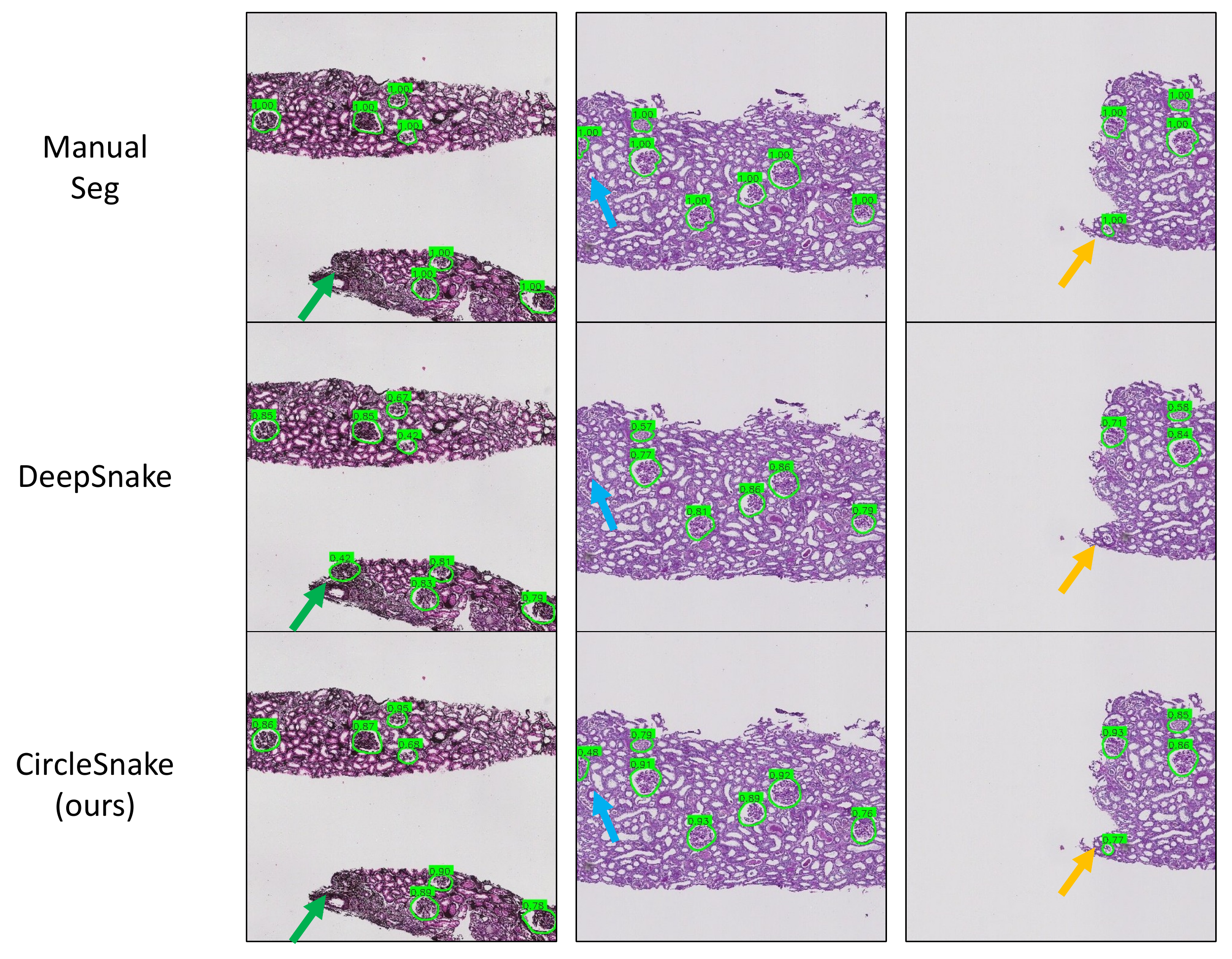}
\end{center}
   \caption{\textbf{Qualitative Results.} This figure shows the qualitative segmentation results of different methods, compared with manual annotations.}
\label{fig:qualitative}
 \end{figure}

\section{Results}
\subsection{Glomeruli Results}
To form a glomeruli dataset, whole slide images of renal biopsies were captured and annotated. The kidney tissue was routinely processed, paraffin embedded, and 3$\mu m$ thickness sections cut and stained with hematoxylin and eosin (HE), periodic acid–Schiff (PAS) or Jones. The  samples were deidentified and these studies were approved by the Institutional Review Board (IRB). The training data contained 704 glomeruli from 42 biopsy samples, the validation data contained 98 glomeruli from 7 biopsy samples, and the testing data contained 147 glomeruli from 7 biopsy samples. Considering the size of a glomerulus~\cite{puelles2011glomerular} and its ratio within an image patch, the original high-resolution whole slide images (0.25 $\mu m$ per pixel) are downsampled to a lower resolution (4 $\mu m$ per pixel). Then, $512 \times 512$ pixels image patches with at least one glomerulus were randomly sampled. These data ultimately comprised a dataset with 7,040 training, 980 validation, and 1,470 testing images.

Referring to Table~\ref{table1}, CircleSnake produces superior detection performane in all average precision metrics. As seen in Table~\ref{table2} and Fig.~\ref{fig:qualitative}, CircleSnake outperforms all baseline methods in Dice score. 
\begin{table}
\caption{Detection and Segmentation Performance on Glomeruli.}
\centering
\begin{tabular}{l@{}c@{\ \ }ccccccc}
\toprule
Methods & Backbone & $AP$ & $AP_{(50)}$ & $AP_{(75)}$ & $AP_{(S)}$ & $AP_{(M)}$ \\
\midrule
CenterNet\cite{zhou2019objects} & DLA & 0.547 & 0.872 & 0.643 & 0.432 & 0.640 \\
CircleNet\cite{nguyen2021circle} & DLA & 0.570 & 0.857 & 0.691 & 0.466 & 0.647 \\
DeepSnake\cite{peng2020deep} (Detection) & DLA & 0.546 & 0.880 & 0.646 & 0.436 & 0.634 \\
CircleSnake (Detection) (Ours) & DLA & 0.590 & 0.890 & 0.701 & \textbf{0.545} & 0.635 \\
DeepSnake\cite{peng2020deep} (Segmentation) & DLA & 0.559 & 0.874 & 0.682 & 0.421 & 0.660  \\
CircleSnake (Segmentation) (Ours) & DLA & \textbf{0.614} & \textbf{0.893} & \textbf{0.737} & 0.544 & \textbf{0.673} \\
\bottomrule
\label{table1}
\end{tabular}
\end{table}

\begin{table}
\caption{Segmentation Performance.}
\centering
\begin{tabular}{l@{}c@{\ \ }cccc}
\toprule
Methods & Backbone & Glomeruli (Dice)\\
\midrule
DeepSnake\cite{peng2020deep} & DLA & 0.804\\
CircleSnake (Ours) & DLA & \textbf{0.849}\\ 
\bottomrule
\label{table2}
\end{tabular}
\end{table}


\section{Conclusion}
In this paper, we introduce CircleSnake, an circle representation based end-to-end deep learning method for instance medical object segmentation. The proposed CircleSnake seamlessly integrated (1) circle detection, (2) circle contour proposal, and (3) circular convolution for segmenting ball-shaped glomeruli, with the superior detectuib instance segmentation performance. The results show that the circle representation does not sacrifice its effectiveness with fewer \ac{DoF} compared with traditional bounding box and octagon representations. While the circle representation has been implemented with the CircleSnake method, it can be extended to any other instance segmentation method.

\textbf{Acknowledgements}:
This work was supported by NIH NIDDK DK56942(ABF).

%
%
\bibliographystyle{splncs04}
\bibliography{main}

\begin{thebibliography}{10}
\providecommand{\url}[1]{\texttt{#1}}
\providecommand{\urlprefix}{URL }
\providecommand{\doi}[1]{https://doi.org/#1}

\bibitem{bueno2020glomerulosclerosis}
Bueno, G., Fernandez-Carrobles, M.M., Gonzalez-Lopez, L., Deniz, O.:
  Glomerulosclerosis identification in whole slide images using semantic
  segmentation. Computer Methods and Programs in Biomedicine  \textbf{184},
  105273 (2020)

\bibitem{d2013rise}
D'Agati, V.D., Mengel, M.: The rise of renal pathology in nephrology: structure
  illuminates function. American journal of kidney diseases  \textbf{61}(6),
  1016--1025 (2013)

\bibitem{gadermayr2017cnn}
Gadermayr, M., Dombrowski, A.K., Klinkhammer, B.M., Boor, P., Merhof, D.: Cnn
  cascades for segmenting whole slide images of the kidney. arXiv preprint
  arXiv:1708.00251  (2017)

\bibitem{ginley2019computational}
Ginley, B., Lutnick, B., Jen, K.Y., Fogo, A.B., Jain, S., Rosenberg, A.,
  Walavalkar, V., Wilding, G., Tomaszewski, J.E., Yacoub, R., et~al.:
  Computational segmentation and classification of diabetic glomerulosclerosis.
  Journal of the American Society of Nephrology  \textbf{30}(10),  1953--1967
  (2019)

\bibitem{he2017mask}
He, K., Gkioxari, G., Doll{\'a}r, P., Girshick, R.: Mask r-cnn. In: Proceedings
  of the IEEE international conference on computer vision. pp. 2961--2969
  (2017)

\bibitem{heckenauer2020real}
Heckenauer, R., Weber, J., Wemmert, C., Feuerhake, F., Hassenforder, M.,
  Muller, P.A., Forestier, G.: Real-time detection of glomeruli in renal
  pathology. In: 2020 IEEE 33rd International Symposium on Computer-Based
  Medical Systems (CBMS). pp. 350--355. IEEE (2020)

\bibitem{huo2021ai}
Huo, Y., Deng, R., Liu, Q., Fogo, A.B., Yang, H.: Ai applications in renal
  pathology. Kidney International  (2021)

\bibitem{kannan2019segmentation}
Kannan, S., Morgan, L.A., Liang, B., Cheung, M.G., Lin, C.Q., Mun, D., Nader,
  R.G., Belghasem, M.E., Henderson, J.M., Francis, J.M., et~al.: Segmentation
  of glomeruli within trichrome images using deep learning. Kidney
  international reports  \textbf{4}(7),  955--962 (2019)

\bibitem{kawazoe2018faster}
Kawazoe, Y., Shimamoto, K., Yamaguchi, R., Shintani-Domoto, Y., Uozaki, H.,
  Fukayama, M., Ohe, K.: Faster r-cnn-based glomerular detection in
  multistained human whole slide images. Journal of Imaging  \textbf{4}(7), ~91
  (2018)

\bibitem{law2018cornernet}
Law, H., Deng, J.: Cornernet: Detecting objects as paired keypoints. In:
  Proceedings of the European Conference on Computer Vision (ECCV). pp.
  734--750 (2018)

\bibitem{lin2017focal}
Lin, T.Y., Goyal, P., Girshick, R., He, K., Doll{\'a}r, P.: Focal loss for
  dense object detection. In: Proceedings of the IEEE international conference
  on computer vision. pp. 2980--2988 (2017)

\bibitem{lin2014microsoft}
Lin, T.Y., Maire, M., Belongie, S., Hays, J., Perona, P., Ramanan, D.,
  Doll{\'a}r, P., Zitnick, C.L.: Microsoft coco: Common objects in context. In:
  European conference on computer vision. pp. 740--755. Springer (2014)

\bibitem{lo2018glomerulus}
Lo, Y.C., Juang, C.F., Chung, I.F., Guo, S.N., Huang, M.L., Wen, M.C., Lin,
  C.J., Lin, H.Y.: Glomerulus detection on light microscopic images of renal
  pathology with the faster r-cnn. In: International Conference on Neural
  Information Processing. pp. 369--377. Springer (2018)

\bibitem{luo2021scpmnet}
Luo, X., Song, T., Wang, G., Chen, J., Chen, Y., Li, K., Metaxas, D., Zhang,
  S.: Scpm-net: An anchor-free 3d lung nodule detection network using sphere
  representation and center points matching. arXiv preprint arXiv:2104.05215
  (2021)

\bibitem{nguyen2021circle}
Nguyen, E.H., Yang, H., Deng, R., Lu, Y., Zhu, Z., Roland, J.T., Lu, L.,
  Landman, B.A., Fogo, A.B., Huo, Y.: Circle representation for medical object
  detection. IEEE Transactions on Medical Imaging  (2021)

\bibitem{peng2020deep}
Peng, S., Jiang, W., Pi, H., Li, X., Bao, H., Zhou, X.: Deep snake for
  real-time instance segmentation. In: Proceedings of the IEEE/CVF Conference
  on Computer Vision and Pattern Recognition. pp. 8533--8542 (2020)

\bibitem{puelles2011glomerular}
Puelles, V.G., Hoy, W.E., Hughson, M.D., Diouf, B., Douglas-Denton, R.N.,
  Bertram, J.F.: Glomerular number and size variability and risk for kidney
  disease. Current opinion in nephrology and hypertension  \textbf{20}(1),
  7--15 (2011)

\bibitem{rehem2021automatic}
Rehem, J.M.C., dos Santos, W.L.C., Duarte, A.A., de~Oliveira, L.R., Angelo,
  M.F.: Automatic glomerulus detection in renal histological images. In:
  Medical Imaging 2021: Digital Pathology. vol. 11603, p. 116030K.
  International Society for Optics and Photonics (2021)

\bibitem{yang2020circlenet}
Yang, H., Deng, R., Lu, Y., Zhu, Z., Chen, Y., Roland, J.T., Lu, L., Landman,
  B.A., Fogo, A.B., Huo, Y.: Circlenet: Anchor-free glomerulus detection with
  circle representation. In: International Conference on Medical Image
  Computing and Computer-Assisted Intervention. pp. 35--44. Springer (2020)

\bibitem{yu2018deep}
Yu, F., Wang, D., Shelhamer, E., Darrell, T.: Deep layer aggregation. In:
  Proceedings of the IEEE conference on computer vision and pattern
  recognition. pp. 2403--2412 (2018)

\bibitem{zhou2019objects}
Zhou, X., Wang, D., Kr{\"a}henb{\"u}hl, P.: Objects as points. arXiv preprint
  arXiv:1904.07850  (2019)

\end{thebibliography}
\end{document}